# AttaCut: A Fast and Accurate Neural Thai Word Segmenter

**Pattarawat Chormai** *
Max Planck School of Cognition
Leipzig, Germany
`pat.chormai@maxplanckschools.de`

**Ponrawee Prasertsom**
Department of Linguistics , Chulalongkorn University
Bangkok, Thailand
`ponrawee.pra@gmail.com`

**Attapol T. Rutherford**
Department of Linguistics, Chulalongkorn University
Bangkok, Thailand
`attapol.t@chula.ac.th`

## Abstract

Word segmentation is a fundamental pre-processing step for Thai Natural Language Processing. The current off-the-shelf solutions are not benchmarked consistently, so it is difficult to compare their trade-offs. We conducted a speed and accuracy comparison of the popular systems on three different domains and found that the state-of-the-art deep learning system is slow and moreover does not use sub-word structures to guide the model. Here, we propose a fast and accurate neural Thai Word Segmenter that uses dilated CNN filters to capture the environment of each character and uses syllable embeddings as features. Our system runs at least $5.6\times$ faster and outperforms the previous state-of-the-art system on some domains. In addition, we develop the first ML-based Thai orthographical syllable segmenter, which yields syllable embeddings to be used as features by the word segmenter.

## 1 Introduction

Word segmentation presents a fundamental challenge for Thai language processing. Many of the downstream natural language processing tasks require that texts be broken into a sequence of words before applying any models. Like Chinese, the Thai script does not mark word boundaries with spaces, so an automatic word segmentation is often required. The Thai script uses 44 consonant symbols, 15 vowel symbols, and 4 tonal symbols. A word is composed of one or more syllables, and each syllable is formed by a set of intricate orthographical rules for valid sequences of Thai alphabet symbols.

Word segmentation is challenging because of linguistic ambiguity and out-of-vocabulary cases. A word can be formed by juxtaposing two "words" e.g. เห็นชอบ (approve) = เห็น (see) + ชอบ (like). This kind of word formation can be detected with a simple dictionary lookup, but harder cases which require context abound in the language. For example, กอดอกไม้ can be segmented into กอ|ดอก|ไม้ (flower bush) or กอด|อก|ไม้ (hugging wooden human chest), but the latter is nonsensical and very unlikely. The local context is needed to select the right segmentation; therefore, dictionaries only provide partial solutions to this problem. A constant stream of new words and loanwords complicates the task of word segmentation further.

---

*Work done while interning at Dr. Attapol T. Rutherford's lab.

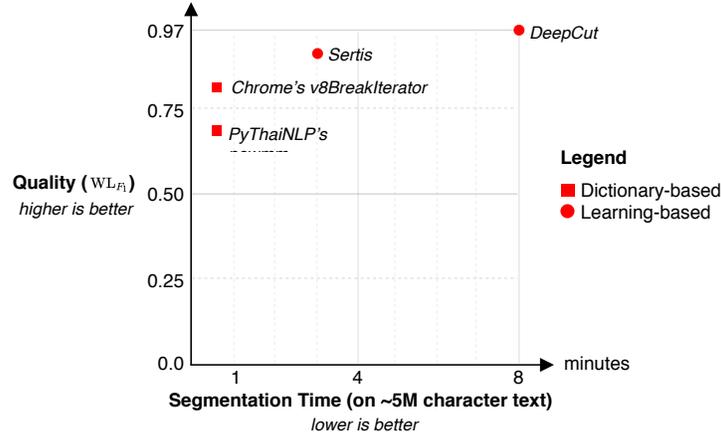

Figure 1: Comparison of segmentation quality and inference time of existing Thai word segmenters. Inference time is measured on a laptop with Intel Core i7 2.2 GHz, 16 GB.

In recent years, a few open-sourced Thai word segmenters have been introduced and used widely in the industry. Notable examples of open-sourced Thai word segmenters include PythaiNLP [16], Sertis [20], and DeepCut [7], which claim good performance according to their own respective benchmarks. The accuracies of these word segmentation systems are not benchmarked on the same datasets for a rigorous comparison within and across domains. We aim to evaluate the speed and accuracy of these previous solutions and compare the trade-offs with our system proposed in this paper.

The popular and powerful Thai word segmenters (DeepCut and Sertis) utilize deep convolutional neural networks (CNNs) or recurrent neural networks (RNNs). The accuracy comes at the high cost of speed, however. The fastest neural model takes 2 minutes to process one million characters on a cloud instance[2]. This rate is unsuitably slow as a first preprocessing step when dealing with a large amount of data or streaming data, where the rate of incoming data exceeds the rate of processing. Figure 1 shows a comparison between speed and segmentation quality of existing Thai word segmenters. In this project, we propose a model that can segment Thai text at least $2\times$ faster than the previous methods.

Our system takes advantage of the fact that every word can be parsed into orthographical syllables. We hypothesize that syllables provide important features for segmentation because word boundaries are a subset of syllable boundaries. We propose a CNN-based word segmentation model that utilizes character and syllable embeddings as the representation and significantly reduces the number layers and computation time needed to achieve the comparable performance.

Our contributions can be summarized as follows:

- We propose a Thai word segmentation model that runs at least $5.6\times$ faster than the previous state-of-the-art system without compromising much segmentation performance.

- We develop the first ML-based Thai orthographical syllable segmenter that resists typos, which actually benefits the performance of our word segmentation model.

- We benchmark the existing word segmentation systems, along with our systems, across multiple data genres to make recommendations for practitioners on speed and accuracy.

---

[2] AWS's t2.medium

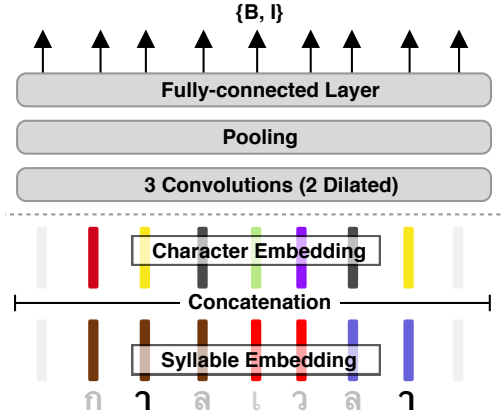

Figure 2: AttaCut-SC: CNN-based word segmenter with character and syllable features. Colors represent different embeddings. Word in figure is กาลเวลา (time). Appendix 7.3.1 provides the details of AttaCut's convolution filters and layers.

## 2 Problem: Thai Word and Syllable Segmentation

Many downstream NLP models require that text is broken down into words or smaller linguistic units. For example, bag-of-word models or RNN models usually require that the text is represented as a sequence of words. A Thai word is the smallest lexical unit that conveys the meaning [2]. การบ้าน (*homework*) is one word, not two, although การ (nominalization prefix) and บ้าน (*house*) are also words in other context. Most ambiguous and debatable cases revolve around the degree of compositionality of nouns and verbs. The meaning of การบ้าน is not composed of การ (nominalization morpheme) and บ้าน (*house*) and therefore should be treated as one word. As a more uncertain example, it is arguable that the meaning of กอดอก (*to cross arms* is composed of กอด (*to hug*) and อก (*breast*) and therefore should not be treated as one word. Our dataset follows the guidelines that favor the segmentation takes the degree of compositionality into account and not grammatical function changes such as nominalization or verbalization.

Syllable segmentation task is defined similarly, but we try to find syllable boundaries instead of word boundaries. It is noteworthy that word boundaries are always subset of syllable boundaries because each syllable belongs to exactly one word. We use this fact as a basis for our model architecture. Unlike word boundaries, syllable boundaries are less ambiguous. An orthographical syllable is defined as a substring in a word that can be pronounced as one or one and a half phonological syllable. For example, a word กอดอก can be segmented into two orthographical syllables กอด ($/\text{god}/$) | อก ($/\text{ok}/$).

## 3 Our Approaches: AttaCut

Our proposed model follows a typical CNN architecture and similar to the state-of-the-art Thai word segmenter, DeepCut. We perform an investigation on DeepCut's architecture, analyzing which parts of the model are important or superfluous: complete details of the analysis can be found at Appendix 7.3.1. We use the insight to design our proposed models that can perform word segmentation quickly while achieving a good level of quality. We employ the dilation technique [23, 27] in the convolution layers, which allows the model to use non-redundant convolution layers that cover sufficient context. Unlike Strubell et al. [23]'s architecture, our convolutional layers have different dilation numbers and kernel widths, and they convolve directly on embeddings. Without hierarchical convolutions, our architecture has less computational dependencies, hence a higher degree of parallelism. In addition, we use syllable embeddings as additional features, which should provide higher-level information than a character type or a character embedding can afford.

Figure 2 shows an overview of our proposed architecture. It is comprised of three one-dimension convolutional layers, pooling layers, and a fully-connected layer. The convolution layers take the concatenation of character and syllable embeddings as input. The concatenation contextualizes the character embedding with its surrounding context provided by the syllable embedding. Therefore, a character will be represented by a different embedding if found in a different syllable. We call this model AttaCut-SC, while our baseline model that use only character embeddings is named AttaCut-C.

We use max pooling to combine extracted features from these convolution layers. After pooling, we have fully-connected layers to derive the probability that the corresponding character is either a starting-word character ($B$) or a in-word character ($I$).

In general, syllable and word segmentation can be formulated as a sequence labelling problem, where we want to assign a label to each position in the sequence: in segmentation problems, the label is binary, representing whether the position is a segment-initial character. A segment here can represent a word or a syllable, depending on the task.

Let $\mathcal{D}$ be a Thai corpus containing Thai sentences $\mathbf{s}_i$ and $\mathbf{y}_i$ represents the segmentation label of $\mathbf{s}_i$. The learning objective is to learn a set of suitable parameters $\hat{\theta}$ of a model $f$ parameterized by $\theta$:

$$\hat{\theta} = \arg\min_{\theta} \mathbb{E}[L(f(\mathbf{s}; \theta), \mathbf{y})] \tag{1}$$

$$\approx \arg\min_{\theta} \frac{1}{|\mathcal{D}|} \sum_{i=1}^{|\mathcal{D}|} L(f(\mathbf{s}_i; \theta), \mathbf{y}_i), \tag{2}$$

where $\mathbb{E}[\cdot]$ denotes expectation and $L$ is a binary cross entropy loss function over all positions in a sequence. We propose a Conditional Random Fields (CRF) syllable segmenter primarily to be used as a preprocessing step for the word segmenter. Historically, automatic syllable segmentation has often been intended for other purposes, such as concatenative speech synthesis [9]. To the best of our knowledge, syllables have never been used as features for ML-based word segmentation systems.

## 4 Experiments

### 4.1 Experiment 1: Syllable Segmenter

We use pycrfsuite implementation of CRF [12, 15] and train the model on Thai National Corpus (TNC) [3]. The TNC contains a subcorpus of 2.56M annotated syllables (around 8M characters). The dataset is split three-way for training, development and testing using the 70:20:10 scheme. The training, development, and test sets contain 1.8M syllables, 0.5M syllables, and 0.25M syllables respectively.

We hypothesize that CRF is suitable for syllable segmentation because of its inclusion of sequential information. We test this hypothesis by comparing it against a maximum entropy model (MaxEnt), trained using the scikit-learn implementation [14]. For both algorithms, we experiment with the following features, with $N$ and window size $W$ of 1 to 4: i) individual characters within $W$ places around on both sides of a potential boundary (Chr); ii) two N-grams on both sides (ChrSpan); iii) N-gram features that include all N-grams within $W$ places on both sides.

Our evaluation employs measures of precision, recall and $F_1$ on two levels: character-level (CL) and syllable-level (SL). CL measures are standard in evaluating word segmentation. Here, they are essentially the same, except for the fact that they are based on that of correct syllable-initial–rather than word-initial–characters. SL measures are calculated based on the number of correct syllables rather than characters. We compute SL precision and recall as follows:

- Syllable-level Precision ($SL_P$) is the ratio between the number of correctly segmented syllables and the number of syllables in prediction;

- Syllable-level Recall ($SL_R$) is the ratio between the number of correctly segmented syllables and the number of syllables in the ground truth.

We use macro-averaging for final statistics.

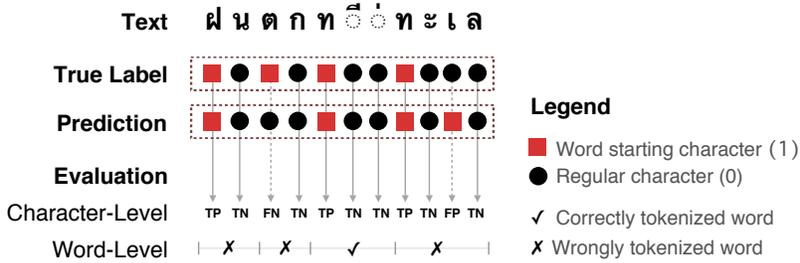

Figure 3: Segmentation evaluation metrics. Correct segmentation is ฝน|ตก|ที่|ทะเล (rain|fall|at|sea).

### 4.2 Experiment 2: Word Segmenter

We use PyTorch [13] for implementation. We train AttaCut word segmenters using Adam [6] and adjust hyperparameters according to training dynamics. We experiment with various layer configurations for both architectures: the best setting for AttaCut-SC contains $158,993$ learning parameters, while AttaCut-C has $173,533$ such parameters.

Our training data is BEST-2010 [11]. Annotated with word boundaries and name entities, the corpus contains 415 Thai documents from four categories: news, articles, encyclopedias, and novels, accounting for 134,107 samples (split by line), around 5.11M words, and 18.74M characters. Balancing the distribution of categories, we take 10% of the training set as a development split. We use the official provided test set (officially named as TEST_100K) for evaluation, which has 2,252 samples, about 128K words, and 496K characters.

Apart from in-domain evaluation on BEST-2010's test set, we also perform cross-domain word segmentation evaluations on another two datasets: i) Thai National Historical Corpus (TNHC) [18] contains 20,791 samples, around 599K words, and 2.14M characters of Thai classical literature documents with word boundaries annotated by humans, and ii) Wisesight corpus [26] contains 26,700 social media messages, labelled in four categories including question, positive, negative, and neutral, with official train and test splits. Because it does not have word boundary annotated, we randomly take 1,000 samples (with 7 spam messages removed) from the test split and manually segment them using a word segmentation standard proposed by Aroonmanakun and others [2]. We call this Wisesight-1000: it has around 22K words, and 75K characters.

We compare our segmenters with: i) PyThaiNLP [16] with its maximal matching engine [22]; ii) Sertis [20], a bidirectional RNN with GRUs with 121K trainable parameters [5, 19]; iii) DeepCut [7], a character-level CNN [28] with a stack of 13 convolutional filters, followed by pooling and fully-connected layers, comprising around 500K trainable parameters. DeepCut's architecture is described in Appendix 7.2.

Evaluating the quality of word segmenters is typically done on a character-level (CL) basis. Standard measured metrics are precision, recall, and $F_1$ of starting-word characters. However, intuitively, when a word is tokenized wrongly, it would consequentially affect the tokenization of following words.

Thus, measuring only the character-level metrics would overestimate the tokenization performance of word tokenizers. Therefore, in this work, we also consider these measures at the word level (WL). As shown in Figure 3, we compute these WL metrics the same way we did with SL metrics, substituting numbers of syllables with numbers of words.

Apart from these quantitative results, we also develop a website[3] that enable qualitative analysis on segmentation results. We refer to Appendix 7.1 for public links to our code that is part of this paper.

Table 1: Syllable segmentation quality (SL$_{F_1}$) on TNC and different methods.

| Algorithm | Features | SL$_{F_1}$ | CL$_{F_1}$ |
|---|---|---|---|
| CRF | Chr (W=4), Trigram (W=4) | **0.96±0.18** | **0.99±0.06** |
| CRF | Chr (W=3), Trigram (W=3) | 0.96±0.18 | 0.99±0.06 |
| CRF | Chr (W=3), ChrSpan (W=3) | 0.95±0.20 | 0.98±0.07 |
| MaxEnt | Chr (W=4), Trigram (W=4) | 0.94±0.22 | 0.98±0.08 |
| MaxEnt | Trigram (W=4) | 0.94±0.22 | 0.98±0.08 |

Table 2: Word segmentation quality (WL$_{F_1}$) on different datasets and methods. (†): Dictionary-based method.

| | Algorithm | | | | |
|---|---|---|---|---|---|
| Dataset | PyThaiNLP† | Sertis | DeepCut | AttaCut-C | AttaCut-SC |
| BEST-2010 | $0.67 \pm 0.19$ | $0.87 \pm 0.16$ | **$0.93 \pm 0.13$** | $0.89 \pm 0.16$ | $0.91 \pm 0.14$ |
| TNHC | **$0.73 \pm 0.21$** | $0.70 \pm 0.23$ | $0.63 \pm 0.26$ | $0.66 \pm 0.24$ | $0.63 \pm 0.26$ |
| Wisesight-1000 | $0.74 \pm 0.21$ | **$0.81 \pm 0.18$** | **$0.81 \pm 0.20$** | $0.80 \pm 0.20$ | **$0.81 \pm 0.20$** |

## 5 Results and Discussion

### 5.1 Syllable Segmentation Performance

We evaluate the quality of syllable segmentation using measures discussed in Section 4.1. As Table 1 shows, CRF models outperform MaxEnt, even with less information. These results suggest that our hypothesis is true, i.e. that sequential information encoded in CRF models matters in syllable segmentation.

Our best model is a CRF model with character and trigram features ($N = 3$, $W = 4$ for both feature types), and the model is further used as part of the word segmentation process.

### 5.2 Word Segmentation Performance

We use measures discussed in Section 4.1 to evaluate word segmentation quality. Table 2 shows that the performance on BEST-2010 of our AttaCut-SC is comparable to DeepCut's performance, with only a few percentage points of WL$_{F_1}$ lower. Although our AttaCut-C achieves slightly lower than DeepCut, its performance serves as a strong baseline for our study: moreover, its performance is higher than the performance of Sertis. The difference between the performances of AttaCut-C and AttaCut-SC verifies our hypothesis that syllable knowledge is important for Thai word segmentation: in fact, with syllable knowledge, AttaCut-SC uses fewer parameters than AttaCut-C.

On cross domain evaluation, DeepCut, Sertis, and AttaCut-SC achieve similar performance on Wisesight-1000. Despite having the lowest word segmenation performance on BEST-2010 and Wisesight-1000, PyThaiNLP is the best word segmenter on TNHC. We refer to Section 5.4 for our explanation about this phenomena.

### 5.3 Speed Benchmark

Because word segmentation is one of the first tasks in NLP pipelines, speed is a key aspect that we have to consider when building a word segmenter. In practical settings, word segmentation is often deployed in distributed and scalable architectures or cloud services. This configuration allows one to operate NLP pipelines at scale while minimizing cost.

Considering these practicalities, we benchmark the speed of the Thai word segmentation models on two cloud instances: AWS's t2.small and t2.medium. We use Wisesight's training set as a benchmark dataset. The dataset contains around 24,063 documents, amounting to around 2.15M characters. Appendix 7.4 discusses our speed benchmark protocol.

---
[3] https://pythainlp.github.io/tokenization-benchmark-visualization/

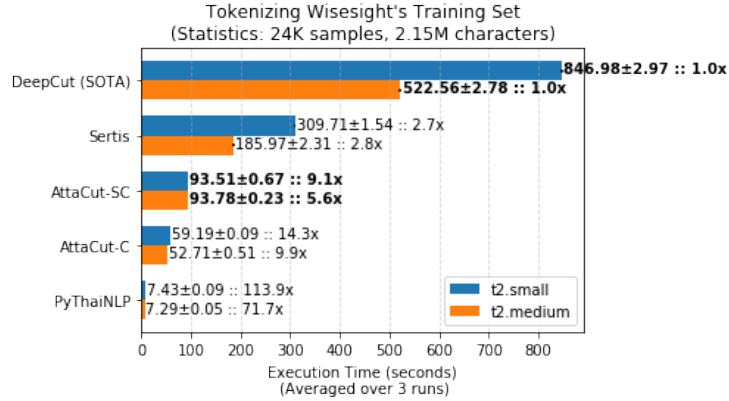

Figure 4: Speed comparison between Thai word segmenters on Wisesight's training set. The factors are relative to DeepCut's corresponding values.

Figure 4 shows that DeepCut, the state of the art, has the longest execution time for segmenting Wisesight's training set. It takes around 846 and 522 seconds on AWS's t2.small and t2.medium respectively. Sertis is the second slowest, taking about 309 and 185 seconds on the two instances, factors of $2.7\times$ and $2.8\times$ faster than DeepCut, respectively. AttaCut-SC completes the segmentation under 100 seconds on both instances, factors of $9.1\times$ and $5.6\times$ faster than DeepCut. AttaCut-C performs the task under 60 seconds ($14.3\times$ and $9.9\times$ faster than DeepCut). PyThaiNLP is the fastest Thai word segmenter, taking less than 10 seconds to achieve the task; however, it suffers from low segmentation quality.

### 5.4 Why do AttaCut and DeepCut fail on TNHC?

Despite displaying state-of-the-art word segmentation results on BEST-2010, both AttaCut and DeepCut models perform poorly on TNHC, while PyThaiNLP performs the best. This poses a questions whether Thai can have a generic word segmenter.

As in other languages, Thai texts can be written in various styles, depending on their purpose. For example, classic literature was typically written with archaic words, or poems were constructed with additional linguistic structures that are quite different from normal texts. This is the case with the TNHC documents: they are classic literature written with poetic techniques.

Sentences in TNHC are composed with words (often short) that are harmonically matched but not semantically similar. Filler words without actual meaning are also used in these documents to create appealing rhythms when read aloud. Therefore, the distribution of words in these documents is completely divergent from our training data (BEST-2010), which contains normal Thai written texts. Because AttaCut and DeepCut are learning-based word segmenters, they segment words based on character correlations that they extract from the training data.

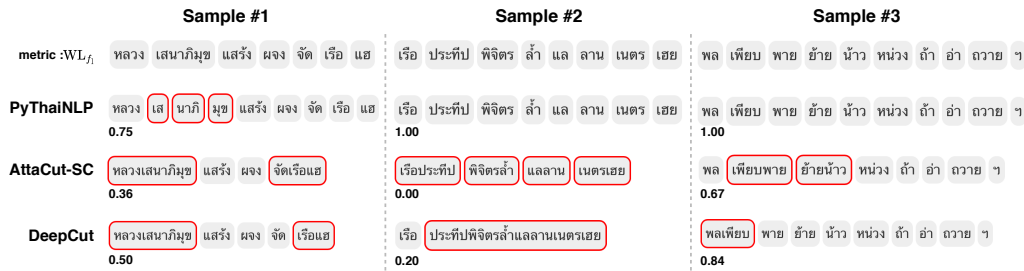

Figure 5: Short words in TNHC causing learning-base word segmenters to fail.

Table 3: PyThaiNLP segments the majority of TNHC samples with a similar number of words, while AttaCut-SC does with fewer.

| TNHC Sample Statistics | | No. Words | | |
|---:|---:|---:|---:|---:|
| Character Length | Percentage | Annotation | PyThaiNLP | AttaCut-SC |
| [0, 20] | **85.50%** | **8.89 ± 3.78** | **8.59 ± 3.71** | 7.22 ± 3.52 |
| (20, 100] | 10.00% | 42.38 ± 20.84 | 36.90 ± 20.01 | 36.24 ± 18.46 |
| (100, 200] | 1.90% | 144.36 ± 30.35 | 129.12 ± 32.49 | 123.98 ± 29.13 |
| (200, 1000] | 2.00% | 420.12 ± 188.15 | 381.94 ± 174.96 | 370.09 ± 169.89 |
| (1000, 2937] | 0.3% | 1517.34 ± 453.75 | 1362.84 ± 439.72 | 1350.54 ± 413.53 |

Figure 5 shows three random samples from TNHC that PyThaiNLP's $WL_{F_1}$ is larger than AttaCut's one by $0.2$: this threshold is set to account for prediction variability. Comparing to annotations, AttaCut-SC and DeepCut attempt to form words from short words, while PyThaiNLP does not. This is a notable case because these formations look reasonable if found in normal Thai texts.

Table 3 shows that PyThaiNLP segments the majority of TNHC samples, whose lengths are fewer than or equal to 20 words, with relatively similar numbers of words, while AttaCut-SC segments them into one word fewer. Although the difference on average is considerably small, using macro-averaging accumulates the large difference in statistics, causing AttaCut-* and DeepCut's $WL_{F_1}$ significantly lower than PyThaiNLP's. Quantitative results from Table 3 support our reasoning that learning-based methods tend to group words together despite compositional irrelevances.

Although we aimed to build a universal word segmenter for Thai, the result of TNHC shows that this might be partly applicable and more experiments should be conducted. One possibility to mitigate the problem is to use transfer learning, training a model on a large corpus before retraining it on the dataset of interest. With that, we provide our implementation with sufficient documentation, such that practitioners can retrain AttaCut models on datasets at hand conveniently. We believe that transfer learning is a reasonable direction that will help alleviate this writing-style problem in Thai word segmentation.

## 6 Related Works

Thai word segmentation has posed its challenges since the digital era. Previous approaches can be categorised into two categories: dictionary-based and learning-based. Dictionary-based approaches rely on an exhaustive dictionary. Poowarawan [17] proposes the first dictionary-based method using a greedy algorithm to decide when a word should be formed. Dictionary-based methods inevitably suffer from unseen words, and hence are harder to generalise to other domains. Sornlertlamvanich [21] proposes an algorithm, called Maximal Matching, to handle such unseen word cases.

Meknavin et al. [10] start formulating word tokenization for Thai as a learning problem. Using hand-crafted features, two algorithms learn to solve ambiguous segmentation cases to aid a dictionary-based segmenter. Theeramunkong et al. [25] present an idea of grouping characters-—Thai Character Clusters (TCCs)—-into a unit that is inseparable based on Thai writing rules, which help reduce chances of segmenting words incorrectly.

Using TCCs, Theeramunkong and Usanavasin [24] develop a decision tree classifier to determine whether a word should be formed from TCCs, based on a predefined metric. Aroonmanakun [1] presents a two-stage word segmentation that incorporates handcrafted syllable features with dynamic programming to form the most reasonable segmentation, while Bheganan et al. [4] use a hidden Markov model to form words that are then verified with a dictionary.

Phatthiyaphaibun and Chaovavanich [16] develops PyThaiNLP, a NLP toolkit for Thai. The package implements various word tokenizers: the default algorithm (newmm) is an extended version of Maximal Matching. Kittinaradorn et al. [7] present DeepCut, a 1-dimension CNN for Thai word segmentation. The state of the art Sertis Co., Ltd. [20] uses a bidirectional-RNN. Lapjaturapit et al. [8] also use a bidirectional RNN to segment words with multiple possible segmentation candidates, which are then selected based on a threshold. However, it is still unclear how one should select the value of the threshold.

# 7 Conclusion

Thai word segmentation is a challenging task in which speed is often exchanged for quality. We proposed an efficient CNN-based word segmenter for Thai that utilizes character and syllable embeddings. The segmenter is at least $5.6\times$ faster than previous state-of-the-art segmenters, and it achieved comparable and, in some domains, better performance. In addition, our analysis shows that learning-based approaches suffer an out-of-domain problem with idiosyncratic datasets such as poetry. Future work could experiment with transfer learning to address this issue.

# Appendix

## 7.1 Reproducibility

Over the course of our project, we have encountered several situations that we wanted to test methods proposed in papers but there was no implementation available, nor concrete details about experiments and datasets. Going through these experiences, we believe that the Thai NLP research community should strive for better reproducibility, paving a common foundation for future advancements. Therefore, we publish every piece of our code and results online, and we strongly encourage other researchers to verify our implementation.

These are sub-projects or resources that are parts of this paper:

- Syllable Segmenter's code
  https://github.com/ponrawee/ssg;

- Word Segmenter's code
  https://github.com/PyThaiNLP/attacut;

- Tokenization Visualization Website
  https://pythainlp.github.io/tokenization-benchmark-visualization/;

- Tokenization Benchmark
  https://www.thainlp.org/pythainlp/docs/dev/api/benchmarks.html;

- AttaCut-SC's training log
  https://www.floydhub.com/pattt/projects/attacut/50;

- AttaCut-C's training log
  https://www.floydhub.com/pattt/projects/attacut/42;

- Thai Word Segmenter Docker containers
  https://github.com/PyThaiNLP/docker-thai-tokenizers.

## 7.2 Why is DeepCut slow?

DeepCut is the state-of-the-art word segmenter for Thai. Despite its high performance on segmentation, using it comes at high computation cost due to a large number of parameters in the model. In particular, DeepCut's structure involves:

- character and character-type embedding layers;
- 1d convolutional layers with kernel widths ranging from 1 to 12;
- pooling layer;
- flattening and concatenation layers;
- fully-connected layer.

The majority of DeepCut's parameters concentrates in the convolutional layers: 415,760 from total 535,025 parameters. Because the kernel widths are varied linearly, the filters of these layers overlap each other; hence, they possibly extract redundant features from the embeddings.

We hypothesise that these redundant convolution layers are DeepCut's computation bottleneck. As shown in Figure 6, we set up an experiment in which we manually disable each of these layers one by one and observe its influence on segmentation quality.

Table 4 shows that DeepCut still performs well when it does not have the convolutional layers with width 7, 9, and 10. Without these three layers, DeepCut becomes 24.60% smaller. Hence, the shrunken DeepCut is approximately 20% faster and can be improved by retraining. This result verifies our hypothesis that some convolution layers of DeepCut are redundant and bring unnecessary computational cost to the model.

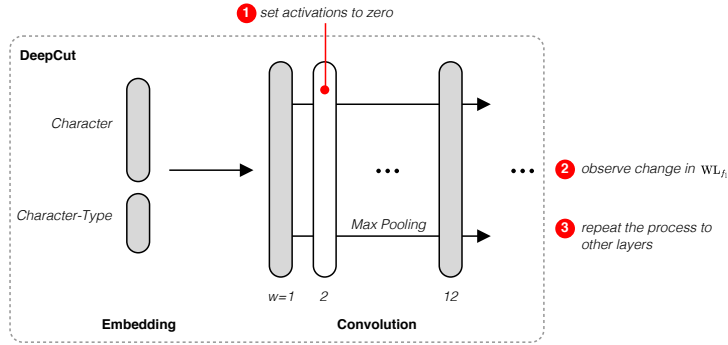

Figure 6: DeepCut Analysis

Table 4: Changes in $WL_{F_1}$ when removing certain layers from DeepCut. Removing the convolutional layers with width 7, 9, and 10 (Conv-7-9-10) affects $WL_{F_1}$ relatively less than other layers, but DeepCut becomes about 24% smaller.

| Inactive Layer | $WL_{F_1}$ | % $\Delta WL_{F_1}$ | Inactive neurons | Size (smaller) |
| --- | --- | --- | --- | --- |
| Original DeepCut | 0.96±0.11 | - | - | - |
| Conv-1 | 0.89±0.21 | -7.29% | 7,605 | 1.42% |
| Conv-2 | 0.90±0.21 | -6.25% | 14,005 | 2.62% |
| Conv-3 | 0.88±0.21 | -8.33% | 20,405 | 3.81% |
| Conv-4 | 0.90±0.21 | -6.25% | 26,805 | 5.01% |
| Conv-5 | 0.90±0.21 | -6.25% | 33,205 | 6.21% |
| Conv-6 | 0.90±0.21 | -6.25% | 39,605 | 7.40% |
| Conv-7 | 0.92±0.21 | -4.17% | 46,005 | 8.60% |
| Conv-8 | 0.91±0.21 | -5.21% | 52,405 | 9.79% |
| Conv-9 | 0.91±0.21 | -5.21% | 44,105 | 8.24% |
| Conv-10 | 0.91±0.21 | -5.21% | 48,905 | 9.14% |
| Conv-11 | 0.90±0.21 | -6.25% | 53,705 | 10.04% |
| Conv-12 | 0.91±0.20 | -5.21% | 38,500 | 7.20% |
| Conv-7-9-10 | 0.90±0.21 | **-6.25**% | 13,1615 | **24.60**% |

### 7.3 AttaCut's Architectures

#### 7.3.1 Convolution Layer Arrangement

Based the results described in Section 7.2, we design a special convolution layer arrangement that allows AttaCut models to extract a similar amount of character features as in DeepCut while minimizing overlaps between filters, using the dilation technique [23, 27]. In total, we use only three convolution layers, i.e. three different kernel widths and dilation numbers. As shown in Figure 7, the configuration contains:

- 1d convolution layer with kernel width 3 and dilation 1[4];
- 1d convolution layer with kernel width 5 and dilation 3;
- 1d convolution layer with kernel width 9 and dilation 2.

Our arrangement of convolution layers cover a context that span 8 characters left and right.

#### 7.3.2 AttaCut-SC

The architecture of AttaCut-SC contains:

- character embedding layer with 32 dimensions;

---
[4]Dilation 1 means no gap in the kernel.

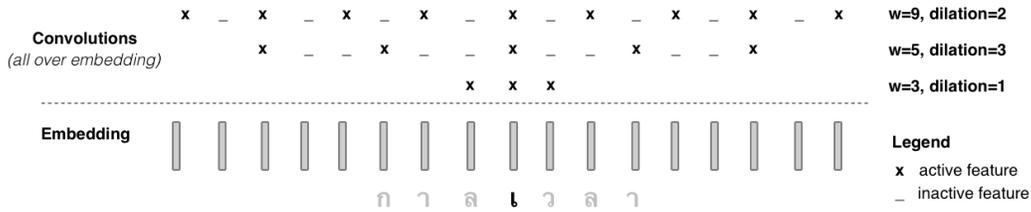

Figure 7: AttaCut's convolution layer arrangement. Three convolutions layers with different kernel widths as well as dilation numbers. Word in figure is กาลเวลา (time).

- syllable embedding layer with 16 dimensions;
- the convolution layers described in Section 7.3 with 64 filters;
- two fully-connected layers with 32 and 1 neurons respectively.

The configuration results in $158,993$ learnable parameters.

### 7.3.3 AttaCut-C

AttaCut-C has a similar layer configuration to AttaCut-SC, except it does not have the syllable embedding layer. Overall, AttaCut-C's architecture includes:

- character embedding layer with 48 dimensions;
- the convolution layers described in Section 7.3 with 196 filters;
- two fully-connected layers with 32 and 1 neurons respectively.

The configuration results in $173,533$ learnable parameters.

### 7.4 Speed Benchmark

We benchmark word segmenters on two cloud instances: t2.small and t2.medium. We perform speed benchmarks on those instances because they are standardized machines and typically used in industry for scalable systems. We use a special OS image provided by AWS, which is highly optimized for numerical computation, such as in neural networks. We develop code[5] that orchestrates benchmarking processes: instantiating a testing instance and installing necessary dependencies. Once the instance is ready, one can access it to run a speed benchmark.

---

[5]https://github.com/heytitle/tokenization-speed-benchmark